\documentclass[]{interact}

\begin{document}

\articletype{FULL PAPER}

\title{Is human face processing a feature- or pattern-based task? Evidence using a unified computational method driven by eye movements}

\author{
\name{C. E. Thomaz\textsuperscript{a},\thanks{CONTACT C. E. Thomaz. Email: cet@fei.edu.br} V. Amaral\textsuperscript{a}, G. A. Giraldi\textsuperscript{b}, D. F. Gillies\textsuperscript{c} and D. Rueckert\textsuperscript{c}}
\affil{\textsuperscript{a}Centro Universitario FEI, Department of Electrical Engineering, Sao Paulo, Brazil\\
       \textsuperscript{b}Laboratorio Nacional de Computacao Cientifica, Rio de Janeiro, Brazil\\
       \textsuperscript{c}Imperial College London, Department of Computing, London, UK}
}

\maketitle

\begin{abstract}
Research on human face processing using eye movements has provided evidence that we recognize face images successfully focusing our visual attention on a few inner facial regions, mainly on the eyes, nose and mouth. To understand how we accomplish this process of coding high-dimensional faces so efficiently, this paper proposes and implements a multivariate extraction method that combines face images variance with human spatial attention maps modeled as feature- and pattern-based information sources. It is based on a unified multidimensional representation of the well-known face-space concept. The spatial attention maps are summary statistics of the eye-tracking fixations of a number of participants and trials to frontal and well-framed face images during separate gender and facial expression recognition tasks. Our experimental results carried out on publicly available face databases have indicated that we might emulate the human extraction system as a pattern-based computational method rather than a feature-based one to properly explain the proficiency of the human system in recognizing visual face information.
\end{abstract}
\begin{keywords}
Human face processing; eye movements; task-driven face dimensions
\end{keywords}

\section{Introduction}

Research on human face processing using eye movements has consistently demonstrated the existence of preferred facial regions or pivotal areas involved in successful identity, gender and facial expression human recognition \cite{hsiao2008,peterson2012,perezmoreno2016,cet2016,bobak2017}. These works have provided evidence that humans analyze faces focusing their visual attention on a few inner facial regions, mainly on the eyes, nose and mouth, and such sparse spatial fixations are not equally distributed and depend on the face perception task under investigation. Faces, therefore, are not only well-known visual stimuli that can be easily recognized by humans \cite{crookes2009}, but also serve as examples of the human ability of extracting relevant visual information very efficiently from the corresponding stimuli to recognize objects of interest with high proficiency.

However, the computational representation and modeling of this apparently natural and heritable embedded human ability \cite{wilmer2010,mckone2010} remains challenging. Although faces are expected to have a global and common spatial layout with all their parts such as eyes, nose and mouth arranged consistently in a multidimensional representation, specific variations in local parts are fundamental to explain our perception of each individual singularity, or groups of individuals when distinguishing, for example, between gender or facial expression \cite{cabeza2000,joseph2003,wallraven2005,maurer2007,shin2011,miellet2011,seo2014}. To understand and emulate how humans accomplish this process of coding faces, it seems necessary to investigate and develop computational extraction methods that explore the combination and relative interaction of the global and local types of information, considering as well the embedded visual knowledge that might be behind the human face perception task under investigation.

In this work, we describe and implement a unified computational method that combines global and local variance information with eye-tracking fixations to represent task-driven dimensions in a multidimensional physiognomic face-space. More importantly, this unified computational method allows the exploration of two distinct embedded knowledge extractions, named here feature- and pattern-based information processing, to disclose some evidence of how humans perceive faces visually. The eye-tracking fixations are based on measuring eye movements of a number of participants and trials to frontal and well-framed face images during separate gender and facial expression classification tasks. In all the automatic classification experiments carried out to evaluate the unified computational method proposed, we have considered: different numbers of face-space dimensions; gender and facial expression sparse spatial fixations; and randomly generated versions of the distribution of the human eye fixations spread across faces. These randomly generated spatial attention maps pose the alternative analysis where there are no preferred viewing positions for human face processing, contrasting the literature findings.

The paper is organized as follows. In section \ref{Materials}, we describe the eye-tracking apparatus, participants, and frontal face stimuli used to generate different fixation images depending on the classification task. Then, section \ref{Method} translates in a unified method the combination of face-space dimensions and eye movements sources of information for feature- and pattern-based multivariate computational analysis. Section \ref{Experiments} describes the eye-tracking experiments carried out and the training and test face samples used from distinct image datasets to evaluate the automatic classification accuracy of the method. All the results have been analyzed in section \ref{Results}. Finally, in section \ref{Conclusion}, we conclude the paper, discussing its main contribution and limitation.

\section{Materials\label{Materials}}

In this section, we describe mainly the eye-tracking apparatus, participants and frontal face stimuli used to generate different fixation images with distinct classification tasks.

\subsection{Apparatus}

Eye movements were recorded with an on-screen Tobii TX300 equipment that comprises an eye tracker unit integrated to the lower part of a 23in TFT monitor. The eye tracker performs binocular tracking at a data sampling rate of 300Hz, and has minimum fixation duration of 60ms and maximum dispersion threshold of 0.5 degrees. These are the eye tracker defaults for cognitive research. A standard keyboard was used to collect participants responses. Calibration, monitoring and data collection were performed as implemented in the Tobii Studio software running on an attached notebook (Core i7, 16Gb RAM and Windows 7).

\subsection{Participants}

A total number of 44 adults (26 males and 18 females) aged from 18 to 50 years participated in this study on a voluntary basis. All participants were undergraduate and graduate students or staff at the university and had normal or corrected to normal vision. Written informed consent was obtained from all participants.

\subsection{Training face database}

\begin{figure}[!htb]
\centering
\includegraphics[width=1.0\linewidth]{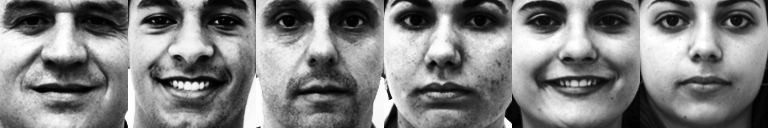}
\caption{A sample of the face stimuli used in this work.}
\label{facesamples}
\end{figure}

Frontal images of the FEI face database \cite{cet2010b} have been used to carry out the eye-movements stimuli. This database contains 400 frontal 2D face images of 200 subjects (100 men and 100 women). Each subject has two frontal images, one with a neutral or non-smiling expression and the other with a smiling facial expression. We have used a rigidly registered version of this database where all these frontal face images were previously aligned using the positions of the eyes as a measure of reference. The registered and cropped images are 128 pixels wide and 128 pixels high and are encoded in gray-scale using 8-bits per pixel. Figure \ref{facesamples} illustrates some of these well-framed images.

\subsection{Stimuli}

Stimuli consisted of 120 frontal face images taken from the training face database. All the stimuli were presented centralized on a black background using the 23in TFT monitor with a screen resolution of 1280x1024 pixels. To improve the stimuli visualization on the TFT monitor all the face images were resized to 512x512 pixels. Presentation of the stimuli was controlled by the Tobii Studio software.

\subsection{Spatial attention maps}

Eye-movements were processed directly from the eye tracker using the Tobii Studio software. Fixation was defined by the standard Tobii fixation filter as two or more consecutive samples falling within a 50-pixel radius. We considered only data from participants for whom on average 25\% or more of their gaze samples were collected by the eye tracker. One participant (1 female) did not meet this criterion and was excluded from the analysis. The standard absolute duration heat maps available at the Tobii Studio software were used to describe the accumulated fixation duration on different locations in the face images at the resolution of 512x512 pixels. These absolute duration heat maps were averaged from all participants and from all face stimuli, generating different fixation images, called here \textit{spatial attention maps}, depending on the classification task.

\section{Method\label{Method}}

The computational method combines face images variance with the spatial attention maps modeled as feature- and pattern-based information sources in a multidimensional representation of the face-space physiognomic dimensions \cite{valentine1991,valentine2015}. It builds on our previous works \cite{cet2010b,cet2012,cet2016} of incorporating task-driven information for a unified multivariate computational analysis of face images using human visual processing.

\subsection{Face-space dimensions}

We have used principal components to specify the face-space physiognomic dimensions \cite{valentine1991,valentine2015} because of their psychological plausibility for understanding the human face image multidimensional representation \cite{sirovich1987,hancock96,todorov11,frowd2015,valentine2015}, where the whole face is perceived as a single entity \cite{rakover02}.

A single entity face image, with $c$ pixels wide and $r$ pixels high, can be described mathematically as a single point in an $n$-dimensional space by concatenating the rows (or columns) of its image matrix \cite{sirovich1987}, where $n = c \times r$. The coordinates of this point describe the values of each pixel of the image and form a $n$-dimensional 1D vector $\mathbf{x}=[x_{1},x_{2},\ldots,x_{n}]^{T}$.

Let an $N \times n$ data matrix $X$ be composed of $N$ face images with $n$ pixels, that is, $X=(\mathbf{x}_{1},\mathbf{x}_{2},\ldots,\mathbf{x}_{N})^{T}$.  This means that each column of matrix $X$ represents the values of a particular pixel all over the $N$ images. Let this data matrix $X$ have covariance matrix
\begin{equation} \label{Scov1}
    S = \frac{1}{(N-1)}\sum_{i=1}^{N}(\mathbf{x}_{i}-\mathbf{\bar{x}})(\mathbf{x}_{i}-\mathbf{\bar{x}})^{T},
\end{equation}
where $\mathbf{\bar{x}}$ is the grand mean vector of $X$ given by
\begin{equation}\label{xmean}
    \mathbf{\bar{x}}=\frac{1}{N}\sum_{i=1}^{N}\mathbf{x}_{i}=[\bar{x}_{1},\bar{x}_{2},\ldots,\bar{x}_{n}]^{T}.
\end{equation}
Let this covariance matrix $S$ have respectively $P$ and $\Lambda$ eigenvector and eigenvalue matrices, that is,
\begin{equation}\label{Ppca}
    P^{T}SP=\Lambda.
\end{equation}
It is a proven result that the set of $m$ ($m \leq n$) eigenvectors of $S$, which corresponds to the $m$ largest eigenvalues, minimizes the mean square reconstruction error over all choices of $m$ orthonormal basis vectors \cite{fukunaga1990}.  Such a set of eigenvectors $P=[\mathbf{p}_{1},\mathbf{p}_{2},\ldots,\mathbf{p}_{m}]$ that defines a new uncorrelated coordinate system for the data matrix $X$ is known as the (standard) principal components.

The calculation of the standard principal components is based entirely on the data matrix $X$ and does not express any domain specific information about the face perception task under investigation. We describe next modifications on this calculation that handle global and local facial differences using feature- and pattern-based combinations of variance and the spatial attention maps.

\subsection{Feature-based combination of variance and spatial attention map (wPCA)\label{wPCA}}

We can rewrite the sample covariance matrix $S$ described in equation (\ref{Scov1}) in order to indicate the spatial association between the $n$ pixels in the $N$ samples as separated $n$ features. When $n$ pixels are observed on each face image, the sample variation can be described by the following sample variance-covariance equation \cite{johnson1998}:
\begin{equation} \label{Scov2}
    S = \left\{s_{jk}\right\} = \left\{\frac{1}{(N-1)}\sum_{i=1}^{N}(x_{ij}-\bar{x}_{j})(x_{ik}-\bar{x}_{k})\right\},
\end{equation}
for $j=1,2,\ldots,n$ and $k=1,2,\ldots,n$. The covariance $s_{jk}$ between the $j^{th}$ and $k^{th}$ pixels reduces to the sample variance when $j=k$, $s_{jk}=s_{kj}$ for all $j$ and $k$, and the covariance matrix $S$ contains $n$ variances and $\frac{1}{2}n(n-1)$ potentially different covariances \cite{johnson1998}.
It is clear from equation (\ref{Scov2}) that each pixel deviation from its mean has the same importance in the standard sample covariance matrix $S$ formulation.

To combine these pixel-by-pixel deviations with the visual information captured by the eye movements, we first represent, analogously to the face images, the corresponding spatial attention map as a $n$-dimensional 1D $\mathbf{w}$ vector, that is,
\begin{equation} \label{w}
    \mathbf{w}=[w_{1},w_{2},\ldots,w_{n}]^{T},
\end{equation}
where $w_{j} \geq 0$ and $\sum_{j=1}^{n}w_{j}=1$. Each $w_{j}$ describes the visual attention power of the $j^{th}$ pixel separately. Thus, when $n$ pixels are observed on $N$ samples, the sample covariance matrix $S^{*}$ can be described by \cite{cet2012,cet2016}
\begin{equation} \label{Rw}
    S^{*} = \left\{s^{*}_{jk}\right\}= \left\{\frac{1}{(N-1)}\sum_{i=1}^{N}\sqrt{w_j}(x_{ij}-\bar{x}_{j})\sqrt{w_k}(x_{ik}-\bar{x}_{k})\right\}.
\end{equation}
It is important to note that $s^{*}_{jk}=s^{*}_{kj}$ for all $j$ and $k$ and consequently the matrix $S^{*}$ is a $n$x$n$ symmetric matrix. Let $S^{*}$ have respectively $P^{*}$ and $\Lambda^{*}$ eigenvector and eigenvalue matrices, as follows:
\begin{equation} \label{Pw}
    P^{*T}S^{*}P^{*}=\Lambda^{*}.
\end{equation}
The set of $m^{*}$ ($m^{*} \leq n$) eigenvectors of $S^{*}$, that is, $P^{*}=[\mathbf{p}^{*}_{1},\mathbf{p}^{*}_{2},\ldots,\mathbf{p}^{*}_{m^{*}}]$, which corresponds to the $m^{*}$ largest eigenvalues, defines the orthonormal coordinate system for the data matrix $X$ called here \textit{feature-based principal components} or, simply, wPCA.

The step-by-step algorithm for calculating these feature-based principal components can be summarized as follows:

\begin{enumerate}
\item Calculate the spatial attention map $\mathbf{w}=[w_{1},w_{2},\ldots,w_{n}]^{T}$ by averaging the fixation locations and durations from face onset from all participants and from all face stimuli for the classification task considered;
\item Normalize $\mathbf{w}$, such that $w_{j} \geq 0$ and $\sum_{j=1}^{n}w_{j}=1$, by replacing $w_{j}$ with $\frac{|w_{j}|}{\sum_{j=1}^{n}|w_{j}|}$;
\item Standardize all the $n$ variables of the data matrix $X$ such that the new variables have $\bar{x}_{j}=0$, for $j=1,2,\ldots,n$. In other words, calculate the grand mean vector as described in Equation (\ref{xmean}) and replace $x_{ij}$ with $z_{ij}$, where $z_{ij}=x_{ij} - \bar{x}_j$ for $i=1,2,\ldots,N$ and $j=1,2,\ldots,n$;
\item Spatially weigh up all the standardized $z_{ij}$ variables using the vector $\mathbf{w}$ calculated in step 2, that is, $z_{ij}^{*}=z_{ij}\sqrt{w_{j}}$;
\item The feature-based principal components $P^{*}$ are then the eigenvectors corresponding to the $m^{*}$ largest eigenvalues of $Z^{*}(Z^{*})^{T}$, where $Z^{*}=\left\{\mathbf{z}_{1}^{*},\mathbf{z}_{2}^{*},\ldots,\mathbf{z}_{N}^{*}\right\}^{T}$ and $m^{*} \leq n$.
\end{enumerate}

\subsection{Pattern-based combination of variance and spatial attention map (dPCA)\label{dPCA}}

We can handle the problem of combining face samples variance with the perceptual processing captured by the eye movements assuming a pattern-based approach rather than a feature-based one as previously described. Here, we would like to investigate the spatial association between the features with their perceptual interaction preserved, not treated separately.

The set of $n$-dimensional eigenvectors $P=[\mathbf{p}_{1},\mathbf{p}_{2},\ldots,\mathbf{p}_{m}]$ is defined in equation (\ref{Ppca}) as the standard principal components, and the $n$-dimensional $\mathbf{w}$ spatial attention representation, where $\mathbf{w}=[w_{1},w_{2},\ldots,w_{n}]^{T}$, is described in equation (\ref{w}).

To determine the perceptual contribution of each standard principal component we can calculate how well these face-space directions align with the corresponding spatial attention map, that is, how well $\mathbf{p}_{1},\mathbf{p}_{2},\ldots,\mathbf{p}_{m}$ align with $\mathbf{w}$, as follows:
\begin{eqnarray}
    k_{1}&=&\mathbf{w}^{T}\cdot \mathbf{p}_{1},  \label{k} \\
    k_{2}&=&\mathbf{w}^{T}\cdot \mathbf{p}_{2},  \nonumber \\
    &&...  \nonumber \\
    k_{m}&=&\mathbf{w}^{T}\cdot \mathbf{p}_{m}.  \nonumber
\end{eqnarray}
Coefficients $k_{i}$, where $i=1,2,\ldots,m$, that are estimated to be $0$ or approximately $0$ have negligible contribution, indicating that the corresponding principal component directions are not relevant. In contrast, largest coefficients (in absolute values) indicate that the corresponding variance directions contribute more and consequently are important to characterize the human perceptual processing.

We select then as the first principal components \cite{cet2010b} the ones with the highest visual attention coefficients, that is,
\begin{equation}\label{dpca_criterion}
    P^{+} = [\mathbf{p}^{+}_{1},\mathbf{p}^{+}_{2},...,\mathbf{p}^{+}_{m^{+}}] = \underset{P}{\arg\max}{\left|P^{T}SP\right|},
\end{equation}
where $\{{\mathbf{p}^{+}_{i}|i=1,2,\ldots\,m^{+}}\}$ is the set of eigenvectors of $S$ corresponding to the largest coefficients $\left\vert k_{1}\right\vert \geq \left\vert k_{2}\right\vert \geq \ldots \geq \left\vert k_{m}\right\vert$ calculated in Equation (\ref{k}), where ($m^{+}<m \leq n$).

The set of $m^{+}$ eigenvectors of $S$, that is, $P^{+}=[\mathbf{p}^{+}_{1},\mathbf{p}^{+}_{2},\ldots,\mathbf{p}^{+}_{m^{+}}]$, defines the orthonormal coordinate system for the data matrix $X$ called here \textit{pattern-based principal components} or, simply, dPCA.

The step-by-step algorithm for calculating these pattern-based principal components can be summarized as follows:

\begin{enumerate}
\item Calculate the spatial attention map $\mathbf{w}=[w_{1},w_{2},\ldots,w_{n}]^{T}$ by averaging the fixation locations and durations from face onset from all participants and from all face stimuli for the classification task considered;
\item Normalize $\mathbf{w}$, such that $w_{j} \geq 0$ and $\sum_{j=1}^{n}w_{j}=1$, by replacing $w_{j}$ with $\frac{|w_{j}|}{\sum_{j=1}^{n}|w_{j}|}$;
\item Calculate the covariance matrix $S$ of the data matrix $X$ and then its respectively $P$ and $\Lambda$ eigenvector and eigenvalue matrices. Retain all the $m$ non-zero eigenvectors $P=[\mathbf{p}_{1},\mathbf{p}_{2},\ldots,\mathbf{p}_{m}]$ of $S$, where $\lambda(j)>0$ for $j=1,2,\ldots,m$ and $m \leq n$;
\item Calculate the spatial attention coefficient of each non-zero eigenvector using the vector $\mathbf{w}$ described in step 2, as follows: $k_{j}=\mathbf{w}^{T}\cdot \mathbf{p}_{j}$, for $j=1,2,\ldots,m$;
\item The pattern-based principal components $P^{+}=[\mathbf{p}^{+}_{1},\mathbf{p}^{+}_{2},\ldots,\mathbf{p}^{+}_{m^{+}}]$, where $m^{+}<m$, are then the eigenvectors of $S$ corresponding to the largest coefficients $\left\vert k_{1}\right\vert \geq \left\vert k_{2}\right\vert \geq \ldots \geq \left\vert k_{m^{+}}\right\vert \geq \ldots \geq \left\vert k_{m}\right\vert$.
\end{enumerate}

\subsection{Geometric Idea}

\begin{figure}[!htb]
\centering
\includegraphics[width=1.0\linewidth]{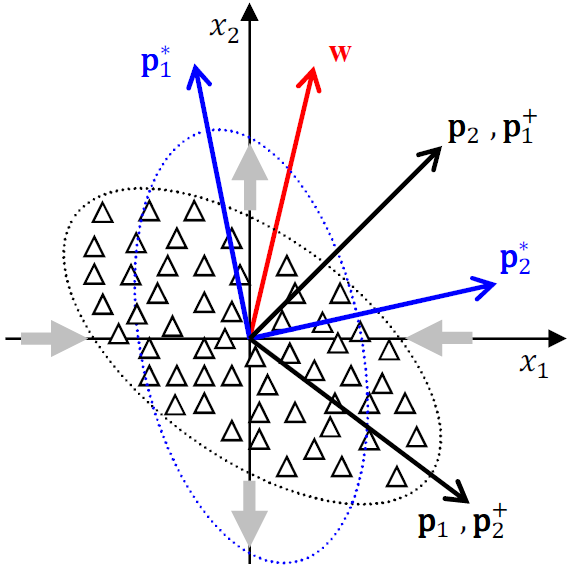}
\caption{An hypothetical example that shows samples (depicted by two-dimensional points represented by triangles) and the geometric idea of the feature- and pattern-based approaches. The former magnifies or shrinks the deviation of each variable separately depending on the direction of $\mathbf{w}$, where $|w_{2}|>|w_{1}|$, and $[\mathbf{p}_{1},\mathbf{p}_{2}]$ and $[\mathbf{p}^{*}_{1},\mathbf{p}^{*}_{2}]$ are respectively the standard and feature-based principal components. The latter re-ranks the standard principal components $[\mathbf{p}_{1},\mathbf{p}_{2}]$ by how well such directions align with $\mathbf{w}$ as a whole, where $|\mathbf{w}^{T}\cdot \mathbf{p}_{2}|>|\mathbf{w}^{T}\cdot \mathbf{p}_{1}|$ and $[\mathbf{p}^{+}_{1},\mathbf{p}^{+}_{2}]$ are consequently the pattern-based principal components.}
\label{wpcaxdpca}
\end{figure}

We show in Figure \ref{wpcaxdpca} the main geometric idea of the feature- and pattern-based principal components. The hypothetical illustration presents samples depicted by triangles along with the spatial attention vector $\mathbf{w}$ represented in red.

It is well known that the standard principal components $[\mathbf{p}_{1},\mathbf{p}_{2}]$ are obtained by rotating the original coordinate axes until they coincide with the axes of the constant density ellipse described by all the samples.

On the one hand, the feature-based approach uses the information provided by $\mathbf{w}$ for each original variable isolated to finding a new orthonormal basis that is not necessarily composed of the same principal components. In other words, in this hypothetical example the influence of the variable deviations on the $x_{2}$ axis will be relatively magnified in comparison with $x_{1}$ because $\mathbf{w}$ is better aligned to the original $x_{2}$ axis than $x_{1}$ one, that is, $|w_{2}|>|w_{1}|$. This is geometrically represented in the figure by large gray arrows, which indicate visually that the constant density ellipse will be expanded in the $x_{2}$ axis and shrunk in the $x_{1}$ axis, changing the original spread of the samples illustrated in black to possibly the constant density ellipse represented in blue. Therefore, the feature-based principal components $[\mathbf{p}^{*}_{1},\mathbf{p}^{*}_{2}]$ would be different from the standard principal components $[\mathbf{p}_{1},\mathbf{p}_{2}]$, because $\mathbf{p}^{*}_{1}$ is expected to be much closer to the $x_{2}$ direction than $x_{1}$, providing a new interpretation of the original data space based on the power of each variable considered separately.

On the other hand, the pattern-based approach uses the information provided by $\mathbf{w}$ as a full two-dimensional pattern, ranking the standard principal components by how well they align with the entire pattern captured by $\mathbf{w}$ across the two-dimensional space samples. Since $\mathbf{w}$ as a whole is better aligned to the second standard principal component $\mathbf{p}_{2}$ direction than the first one $\mathbf{p}_{1}$, i.e. $|\mathbf{w}^{T}\cdot \mathbf{p}_{2}|>|\mathbf{w}^{T}\cdot \mathbf{p}_{1}|$, then $\mathbf{p}^{+}_{1}=\mathbf{p}_{2}$ and $\mathbf{p}^{+}_{2}=\mathbf{p}_{1}$. Therefore, the pattern-based approach selects as its first principal component the standard variance direction that is most efficient for describing the whole spatial attention map, which comprises here the entire pattern of the eye movements across the whole faces, rather than representing all the pixels visual attention power as unit apart features.

%-------------------------------------------------------------------------
\section{Experiments\label{Experiments}}

The experiments consisted of two separate and distinct classification tasks: (1) gender (male versus female) and (2) facial expression (smiling versus neutral). During the gender experiments, 60 faces equally distributed among gender were shown on the Tobii eye tracker, all with neutral facial expression (30 males and 30 females). For the facial expression experiments, all the 60 faces shown were equally distributed among gender and facial expression (15 males smiling, 15 females smiling, 15 males with neutral expression and 15 females with neutral expression).

Participants were seated in front of the eye tracker at a distance of 60cm and initially filled out an on-screen questionnaire about their gender, ethnicity, age and motor predominance. They were then instructed to classify the faces using their index fingers to press the corresponding two keys on the keyboard. Participants were asked to respond as accurately as possible and informed that there was no time limit for their responses. Each task began with a calibration procedure as implemented in the Tobii Studio software. On each trial, a central fixation cross was presented for 1 second followed by a face randomly selected for the corresponding gender or facial expression experimental samples. The face stimulus was presented for 3 seconds in both tasks and was followed by a question on a new screen that required a response in relation to the experiment, that is, "Is it a face of a (m)ale or (f)emale subject?" or "Is it a face of a (s)miling or (n)eutral facial expression subject?". Each response was subsequently followed by the central fixation cross, which preceded the next face stimulus until all the 60 faces were presented for each classification task. Each participant completed 60 trials for the gender classification task and 60 trials for the facial expression one with a short break in between the tasks.

We adopted a 10-fold cross validation method drawn at random from the gender and smiling corresponding sample groups to evaluate the automatic classification accuracy of the feature- and pattern-based dimensions. Additionally to the FEI face samples described previously and used for training only, we have used frontal face images of the well-known FERET database \cite{pwh98}, registered analogously to the FEI samples, for testing. In the FERET database, we have also considered 200 subjects (107 men and 93 women) and each subject has two frontal images (one with a neutral or non-smiling expression and the other with a smiling facial expression), providing a total of two distinct training and test sets of 400 images to perform the gender and expression automatic classification experiments. We have assumed that the prior probabilities and misclassification costs are equal for both groups. On the principal components subspace, the standard sample group mean of each class has been calculated from the corresponding training images and the minimum Mahalanobis distance from each class mean \cite{fukunaga1990} has been used to assign a test observation to either the smiling and non-smiling classes in the facial expression experiment, or to either the male or female classes in the gender experiment.

In all the automatic classification experiments, we have considered different numbers of principal components, both gender and facial expression task-driven spatial attention maps accordingly, and their corresponding randomly generated versions with the distribution of the human eye fixations uniformly spread across faces to pose the alternative analysis where there are no preferred viewing positions for human face processing.

\section{Results\label{Results}}

The classification results of the 43 participants on the gender and facial expression tasks were all above the chance level (50\%). Their performance on the male versus female (gender) and smiling versus non-smiling (facial expression) tasks were on average $97.2\%(\pm2.3\%)$ and $92.8\%(\pm4.3\%)$, respectively.

\begin{figure}[!htb]
\centering
\includegraphics[width=1.0\linewidth]{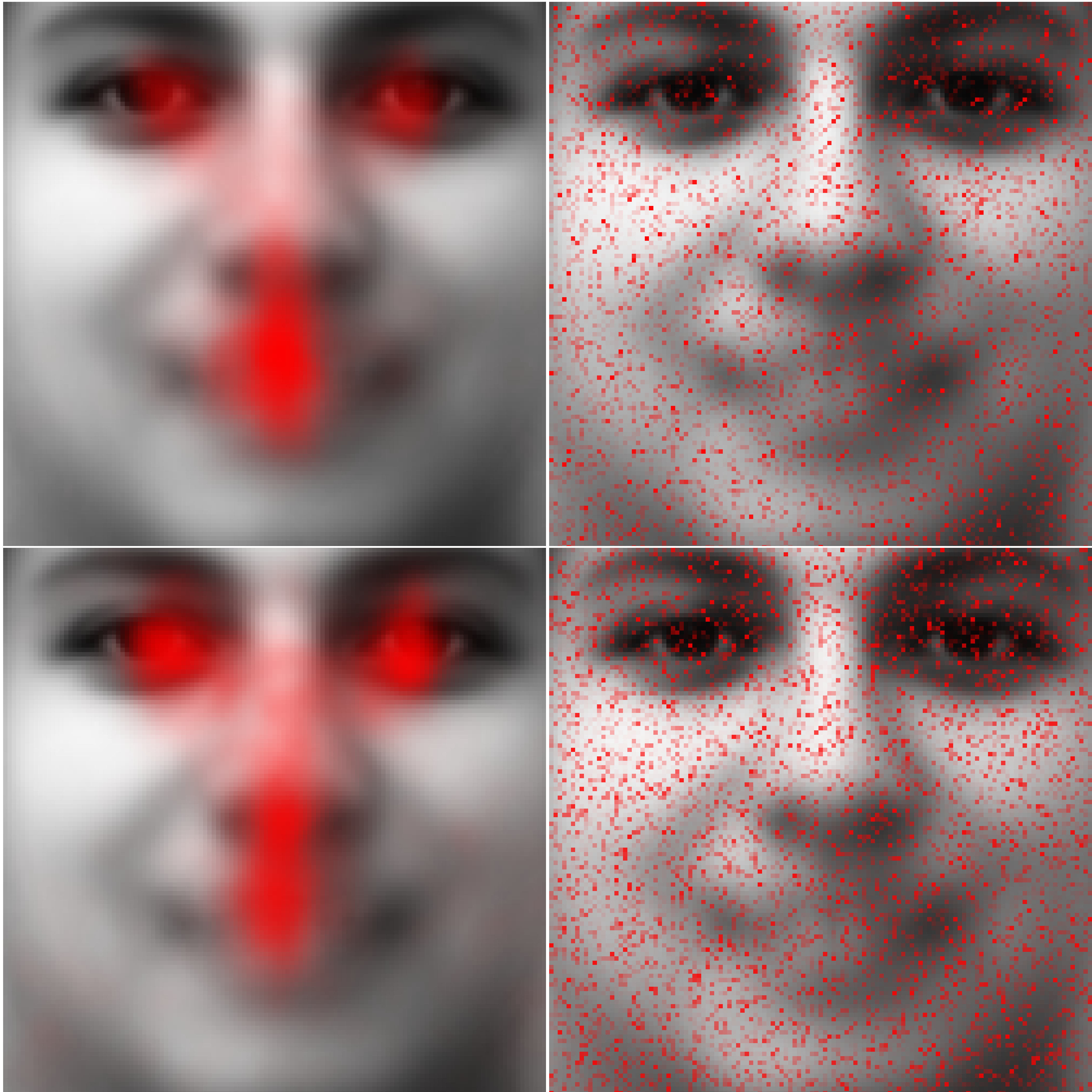}
\caption{An illustration of the spatial attention maps (left) and their corresponding randomly generated versions (right). The upper and lower panels describe the facial expression and gender classification tasks, respectively, superimposed on the grand mean face of the training database used.}
\label{fixations}
\end{figure}

Figure \ref{fixations} illustrates the spatial attention maps (left side) along with their corresponding randomly generated versions (right side) used to calculate the feature- and pattern-based principal components. The spatial attention maps (left) are summary statistics that describe the central tendency of the fixation locations and durations from face onset from all participants and from all face stimuli after 3 seconds for the facial expression (top panel) and gender (bottom) classification tasks. We have disregarded the first two fixations of all participants to avoid the central cross bias. There are location similarities in the manner all the faces were perceived, highlighting relevant proportion of fixations directed at mainly the pivotal areas of both eyes, nose and mouth. These results show that participants made slightly different fixations in the two classification tasks on these areas known as optimal for human processing of the entire faces \cite{hsiao2008,peterson2012,bobak2017}. In contrast, as expected, the randomly generated versions (right) of the spatial attention maps are uniformly distributed, showing essentially a sub-sampling of the entire face without any preferred features or viewing positions.

\begin{figure}[!htb]
\centering
\includegraphics[width=0.9\linewidth]{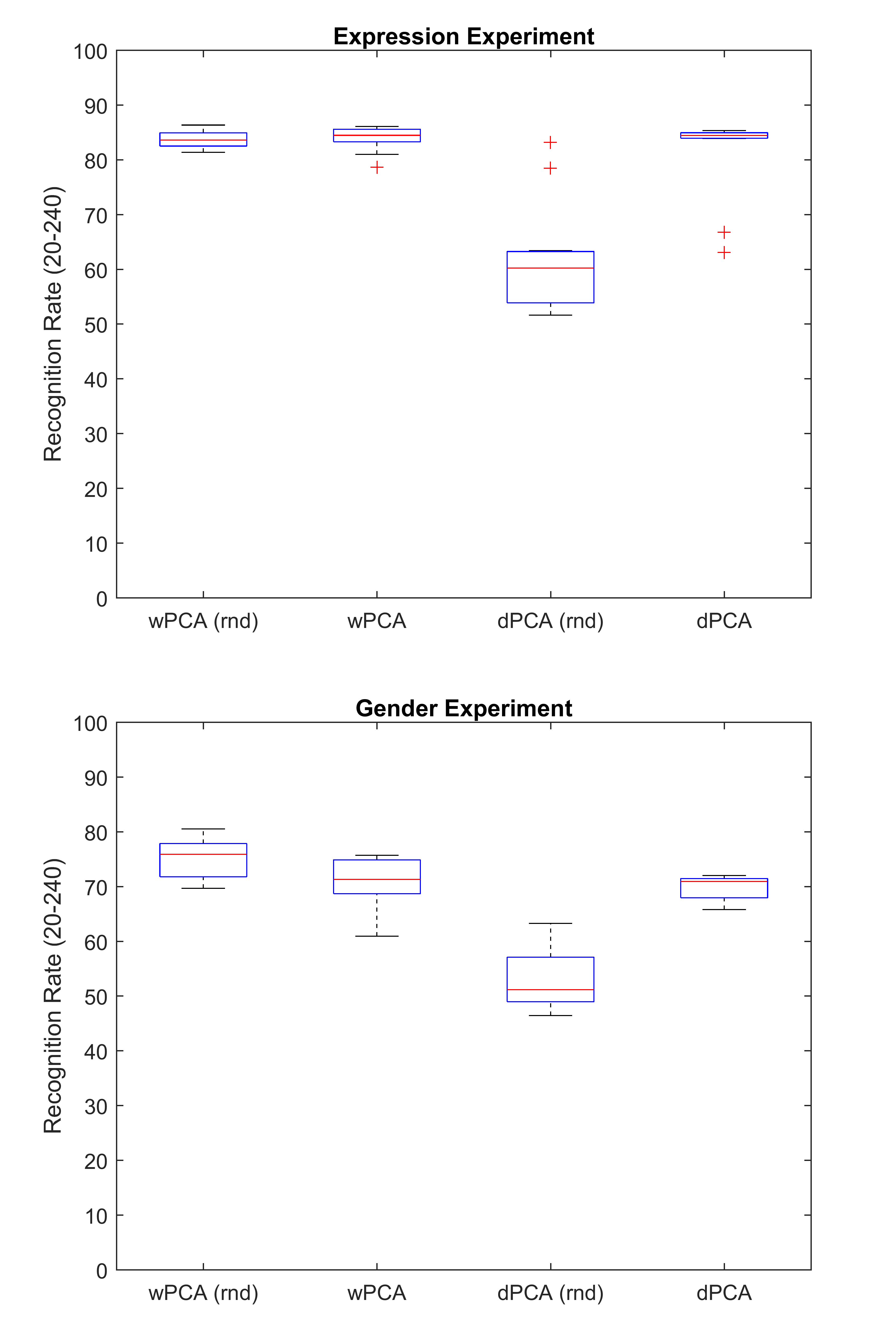}
\caption{Facial expression (top) and gender (bottom) boxplots of the recognition rates of feature-based (wPCA) and pattern-based (dPCA) dimensions given the corresponding spatial attention maps and their randomly generated versions (rnd). The number of principal components considered for automatic classification varied from 20 to 240.}
\label{boxplots}
\end{figure}

Figure \ref{boxplots} shows the recognition rates of the facial expression (top panel) and gender experiments (bottom panel) for the feature- and pattern-based principal components using the spatial attention maps and their corresponding random versions. The number of principal components considered varied from 20 to 240 because all the recognition rates leveled off or decreased with additional components. We can see that both feature- and pattern-based automatic mappings of the high-dimensional face images into lower-dimensional spaces are accurately equivalent, with no significant statistical difference on their recognition rates, when using the facial expression and gender spatial attention maps to highlight accordingly the preferred inner face regions for automatic classification. In other words, both processings have shown to be computationally effective to automatically classify the facial expression and gender samples used.

However, the feature-based behavior is noteworthy when using the random versions of the aforementioned maps. In Figure \ref{boxplots}, there is no statistical difference between the feature-based and its random version of results on the facial expression experiments and, in fact, there is some statistical significant improvement ($p<0.05$) on its automatic recognition performance when using the randomly generated version of the spatial attention maps in the gender experiments. On the condition of analyzing frontal and well-framed face images, the results indicate that the feature-based approach shows no specific exploitation of the manner in which all the participants have viewed the entire faces when classifying the samples. These results suggest indeed that there is no critical region or pivotal areas involved in successful gender and facial expression human recognition, despite the clear evidence of focused visual attention on the eyes, nose and mouth described in the previous Figure \ref{fixations}.

Interestingly, though, the findings of the pattern-based dimensions are exactly the opposite. We can see clearly in Figure \ref{boxplots} statistical differences between the pattern-based and its random version results ($p<0.001$) on both facial expression and gender experiments, where the preferred participants fixation positions augment considerably the automatic recognition of their face-space dimensions when using the spatial attention maps rather than their randomly generated versions. These results provide multivariate statistical evidence that faces seem to be analysed visually using a pattern-based strategy, instead of decomposing such information processing into separate and discrete local features.

\section{Conclusion\label{Conclusion}}

This work provides theoretical and empirical evidence about the processes underlying human face visual cognition using frontal and well-framed face images as stimuli. Exploring eye movements of a number of participants on gender and facial expression distinct classification tasks, we have been able to implement an automatic multivariate statistical extraction method that combines variance information with sparse visual processing about the task-driven experiment.

Our experimental results carried out on publicly available face databases have suggested that the proficiency of the human face processing can be explained computationally as a pattern-based feature extraction process, but not as a feature-by-feature one. The pattern-based computation of the spatial association between the face image features with their visual perceptual interaction preserved has shown statistically significant differences between the preferred human eye fixations and their randomly generated versions, confirming systematically the literature findings of few but sufficient viewing positions for successful human face recognition. Conversely, the feature-based computational results suggest that there is no critical region or pivotal areas involved in such processing, and a random sub-sampling of the entire face without any preferred features or viewing positions provides comparable accuracies, despite the well-known evidence, shown in this work as well, for focused human visual attention on the eyes, nose and mouth. In short, to convey the visual cues that might create a perceptually plausible approach for coding face images for recognition, the results described here indicate that we might emulate the human extraction system as a pattern-based computational method rather than a feature-based one to advance the development of more efficient methods for the spatial analysis of visual face information.

All the analyses presented here are based on the concept of psychologically plausible face-space dimensions described as principal components of high-dimensional image vectors. Although this concept has become established as an effective linear model for representing the dimensions of variation that occur in collections of human faces \cite{valentine1991,sirovich1987,hancock96,rakover02,todorov11,frowd2015,valentine2015}, it needs to reshape the 2D face input images into 1D vectors, breaking the natural structure of the human face perception. Further work on multilinear subspace learning might result in more compact representation of the human face processing, preserving the original data structure and leading perhaps to a more effective computational approach for coding face images when reducing the dimensionality of the high-dimensional image inputs.

\section*{Acknowledgement(s)}

The authors would like to thank the financial support provided by FAPESP (2012/22377-6) and CNPq (309532/2014-0). This work is part of the Newton Advanced Fellowship appointed by the Royal Society to Carlos E. Thomaz.

\bibliographystyle{tfnlm}
\bibliography{mybib}

\end{document}